\documentclass[sigconf]{aamas}  

\usepackage{booktabs}
\usepackage{algorithm,algpseudocode}
\usepackage{amsthm}
\usepackage{subcaption}
\usepackage{makecell}
\usepackage{balance}

\newcommand{\alg}{\textsc{PLOTS}}
\newcommand{\algi}{\textsc{PLOTS-Sketch}}
\newcommand{\algii}{\textsc{PLOTS-NoSketch}}
\newcommand{\algbase}{\textsc{BPS}}
\newcommand{\algbaselong}{\textsc{Backtracking Procedure Search (BPS)}}
\newcommand{\algoracle}{\textsc{BPSOSA}}
\newcommand{\algoraclelong}{\textsc{BPS with Oracle Sketch Alignment (BPSOSA)}}

\setcopyright{ifaamas}  
\acmDOI{}  
\acmISBN{}  
\acmConference[AAMAS'19]{Proc.\@ of the 18th International Conference on Autonomous Agents and Multiagent Systems (AAMAS 2019)}{May 13--17, 2019}{Montreal, Canada}{N.~Agmon, M.~E.~Taylor, E.~Elkind, M.~Veloso (eds.)}  
\acmYear{2019}  
\copyrightyear{2019}  

\begin{document}

\title{PLOTS: Procedure Learning from Observations using Subtask Structure}  




%
\author{Tong Mu}
\affiliation{%
 \institution{Department of Electrical Engineering}
 \institution{Stanford University}
}
\email{tongm@stanford.edu}

\author{Karan Goel}
\affiliation{%
 \institution{Department of Computer Science}
 \institution{Stanford University}
}
\email{kgoel@cs.stanford.edu}

\author{Emma Brunskill}
\affiliation{%
 \institution{Department of Computer Science}
 \institution{Stanford University}
}
\email{ebrun@cs.stanford.edu}

\begin{abstract}  


In many cases an intelligent agent may want to learn how to mimic a single  observed demonstrated trajectory. In this work we consider how to perform such procedural learning from observation, which could help to enable agents to better use the enormous set of video data on observation sequences. Our approach exploits the properties of this setting to incrementally build an open loop action plan that can yield the desired subsequence, and can be used in both Markov and partially observable Markov domains. In addition, procedures commonly involve repeated extended temporal action subsequences. 
Our method optimistically explores actions to leverage potential repeated structure in the procedure. In comparing to some state-of-the-art approaches we find that our explicit procedural learning from observation method is about 100 times faster than policy-gradient based approaches that learn a stochastic policy and is faster than model based approaches as well. We also find that performing  optimistic action selection yields substantial speed ups when latent dynamical structure is present.

\end{abstract}

%


\maketitle

\section{INTRODUCTION}


An incredible feature of human intelligence is our ability to imitate behavior, such as a procedure, simply by observing it. Whether watching someone perform CPR or observing a chef cook an omelette, people can learn to mimic such demonstrations with relative ease. While there has been extensive interest in learning from demonstration, particularly for robotics, this work typically assumes access to the demonstrator's actions and resulting impacts on the environment 
(observations). In contrast, there exists orders of magnitudes more 
demonstration data that only contains the observation trajectories but not the actions -- we see the result of the motor commands when cracking an egg, but not the motor commands themselves. 

In this paper we focus on how an agent can efficiently learn 
to match a single observation sequence, which 
we call \textit{procedure learning from observation}. The agent has access to a simulator of the environment, and must efficiently learn to match the demonstrated behavior. For this to be possible, the dynamics of the underlying domain must be deterministic, at least at the 
level of the observation sequence\footnote{In other words, there must exist 
at least one single sequence that can deterministically achieve the 
demonstrated observation trajectory.}. There are many cases in which we would 
like an agent to perform such procedure learning from a single demonstration -- \emph{e.g.} to learn a recipe, play a musical piece, swing a golf club or fold a shirt. 
Often, such procedures themselves involve repeated substructure in the necessary action sequence, where subsequences of actions are repeated several times. Cracking a series of eggs to make an omelette or performing multiple rounds of chest compressions during CPR are examples of this. Our algorithm leverages the structure of such settings to improve the data-efficiency of an agent learning to mimic the desired observation sequence. 

Procedure learning from a single demonstrated observation trajectory 
relates to two recent research threads. The prior work on learning from observations~\cite{IJCAI2018-torabi,goo2018learning,sermanet2016unsupervised,liu2017imitation} has focused on learning  generalizable conditional policies. In contrast we focus on building 
an open-loop action plan (which must be sufficient to enable optimal behavior 
in procedural imitation), and find this can drastically reduce the amount 
of experience needed for an agent to learn. Other work has sought to 
leverage provided 
policy sketches~\cite{Andreas2017ModularMR}, weak supervision of structure in the decision policy, in order to speed and/or improve learning in the multi-task reinforcement learning and imitation learning (with provided actions) setting~\cite{Andreas2017ModularMR,shiarlis2018taco}. In this work we consider 
how similar policy sketches can be used in the observational learning setting, 
and, unlike prior related work, our focus is particularly on inferring 
or assuming such structure in order to speed learning of the procedure. Unlike some related work~\cite{IJCAI2018-torabi}, our work does not assume the observation 
space is Markov and it can be applied to domains with perceptual aliasing 
in the observation space. 

Our two key ideas are to learn a plan rather than a policy, and to opportunistically drive action selection to leverage potential repeated structure in the 
procedure. To do so we introduce a method loosely inspired by backtracking 
beam search. Our method incrementally constructs a partial plan to yield observations that match the first part of the demonstrated observation 
sequence. To achieve this, it maintains a set of possible clusterings or alignments of the actions in that plan, using these to guide exploration to mimic the remaining part of the demonstration. 

We find that our algorithm learns substantially 
faster than policy gradient approaches in both Markov and 
non-Markov simulated, deterministic domains. We find additional benefits 
from leveraging additional information in the form of input policy sketches. 
Interestingly we also find that these benefits can 
be obtained even when such policy sketches are \emph{not} provided, 
by a variant of our algorithm that 
opportunistically biases exploration towards potential repeated action 
sub-sequences. We conclude with a brief investigation of how our approach 
may be useful in a continuous domain, and a discussion of limitations 
and future directions.

\section{RELATED WORK}

Procedure learning from observation is related to many ideas, drawing on insights from the extensive learning from demonstration literature and hierarchical learning.
\subsection{Learning from Observation}
Inspired by human learning from direction observation (without access to the actions), learning by observation has attracted increasing interest in the last few years \cite{stadieThirdPerson2017,duan2017one,liu2017imitation,IJCAI2018-torabi,yu2018one,goo2018learning,sermanet2016unsupervised}. Observational learning has the potential to allow humans or artificial agents to learn directly from raw video demonstrations. Due to the wealth of such recorded videos, successful observational learning could enable important advances in agent learning. Observational learning can potentially enable a learner to achieve the task with an entirely different set of actions than the original demonstrator (\emph{e.g.} robotic manipulator vs human hands) and to translate shifts in the observation space between the demonstrator and the learner (a new viewpoint, a different background, \emph{etc}). Several papers have focused on robotic learning from third-person demonstrations, particularly where the viewpoint of the demonstrator is different from the target context~\cite{stadieThirdPerson2017,duan2017one}. Unlike such work we focus on the simpler case where the agent's observation space matches the demonstrator's observation space; at least on the subset of features required to specify the reward (\emph{e.g.} for playing the piano, the visual features might not match but the audio features must). However, our work also tackles the harder case of learning from only a single demonstration. Prior work that operates on a single observational demonstration typically assumes some additional experience or training -- such as prior experience used to learn a dynamics model for the learner's environment~\cite{IJCAI2018-torabi}, a set of expert demonstrations in different contexts~\cite{liu2017imitation}, a batch of prior paired expert demonstrations and robot demonstrations (complete with the robot actions)~\cite{yu2018one} -- that can then speed agent learning given a new observation sequence. Such learning of transferable (dynamics or policy) models is often framed as a form of meta-learning or transfer learning~\cite{finn2017one}, and has led to exciting successes for one-shot imitation learning in new tasks. 

In contrast, our focus in this paper is in enabling fast procedure learning from a single observation trajectory -- learning to exactly mimic the trajectory as performed by the demonstrator. If the domain has stochastic dynamics, this is in general impossible, so we focus on the case where the dynamics are deterministic,at least at the level of the observations. We highlight this point because the observations may already be provided at some level of perceptual abstraction rather than low-level sensor readings. For example, the motion of a robot may be slightly jittery, but we can define extended temporal action sequences that can deterministically transition between high-level, abstract observations, like whether the agent is inside or outside a building.

\subsection{Leveraging Weak Hierarchy Supervision}
A number of papers have considered hierarchical imitation learning. The majority of such work assumes the agent has access to demonstrated state-action trajectories, where behavioral 
cloning could be applied, but no additional supervision (though exceptions which leverage additional expert interactions exist \emph{e.g.}~\cite{abdo2013learning}). The agent performs unsupervised segmentation or sub-policy discovery from the observed demonstration trajectories~\cite{schaal2006dynamic,ekvall2006learning} using (for example) changepoint detection \cite{konidaris2012robot}, latent temporal variable modeling (e.g.~\cite{niekum2015learning}), expectation-gradient approaches (e.g. \cite{bui2002policy,daniel2016probabilistic,pmlr-v78-krishnan17a}) or mixture-of-experts modeling~\cite{henderson2017optiongan}. Such methods often leverage parametric assumptions about the underlying domain to help guide the discovery of hierarchical structure. Often, the learned sub-policies have been shown to accelerate the learner on the same tasks (as demonstrated) and/or benefit transfer learning to related tasks. A related idea involves inferring a sequence of abstracted actions (named a workflow) consistent with a demonstration: there can be multiple potential workflows per demonstration~\cite{liu2018reinforcement}. The workflow structures are used to prioritize exploration for related webpage tasks, and show promising improvements, but the workflow inference presupposes particular properties of webpage tasks. In contrast to such unsupervised option discovery, recent work~\cite{shiarlis2018taco} shows that if the demonstrated state-action trajectories are weakly labeled with the sequence of subtasks required to complete the task (inspired by the labels provided in modular policy sketches~\cite{Andreas2017ModularMR}), this can yield performance almost as strong as if full supervision of the sub-policy segmentation is provided, though the authors did not compare to unsupervised option-discovery methods. 

Our work also seeks to leverage such policy sketches in imitation learning, but, to the best of our knowledge, in contrast to the above hierarchical imitation learning research, our work is the first to consider hierarchical learning from observation. 


\subsection{Additional Related Work}
Procedural learning from observations is possible when the domain is 
deterministic. Prior work has shown stronger performance guarantees when the decision process is deterministic~\cite{Wen17} compared 
to more general, stochastic decision processes. Intuitively, deterministic 
domains imply that an open loop action or plan is optimal, compared to a 
state-dependent policy. We find similar performance benefits in our 
setting. Many tasks are deterministic or can be approximated as such by employing the right observation abstraction. 

Our technical approach for performing procedural learning by observation is related to backtracking beam search \cite{zhou2005beam}. Backtracking beam search is a strategy for exploring graphs 
efficiently by only exploring a fixed number $b$ of the most promising next nodes at each time-step while maintaining a stack of unexplored nodes to backtrack to, guaranteeing correctness. 




\section{SETTING} \label{setting}
\begin{figure*}
\centering
  \begin{minipage}[c]{0.7\textwidth}
    \includegraphics[width=\textwidth]{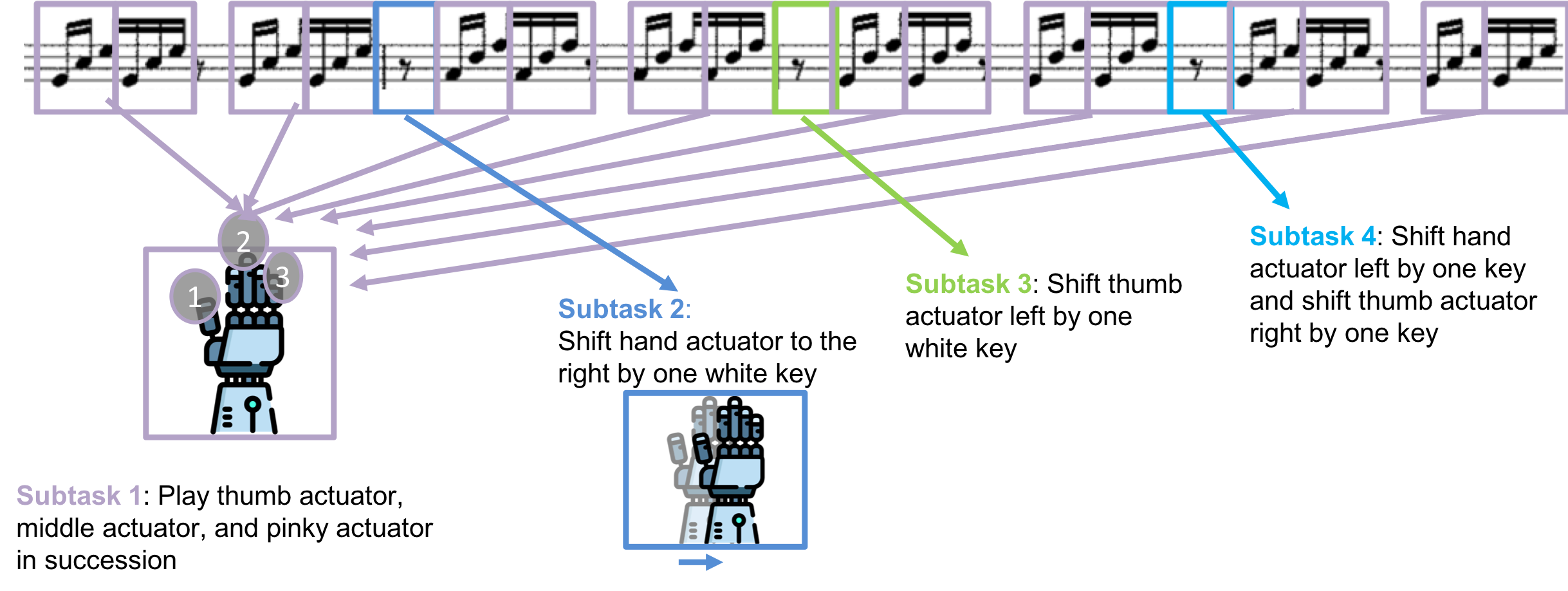}
  \end{minipage}\hfill
  \begin{minipage}[c]{0.25\textwidth}
    \caption{An illustration of our Piano domain inspired by Bach's Prelude in C with the subtasks labeled. Icons used to make this image are credited in~\cite{image_credits}
    } \label{robot_piano}
  \end{minipage}
\end{figure*}
We define the task of procedure learning from observation as: 
given a single fixed 
input observation sequence $\mathcal{Z}^*=(z^*_1,z^*_2,\ldots,z^*_H)$ 
, the agent must learn an action plan  $a^*_{1:H}=(a^*_1,a^*_2,\ldots,a^*_H)$ that, when executed, yields the 
same observation sequence $\mathcal{Z}^*$.  Specifically 
we assume the agent is acting in a stationary, deterministic, potentially 
partially observable Markov decision 
process consisting of a fixed set of actions $A$, states $S$ and observations $Z.$ 
The dynamics model is deterministic: for each $(s,a)$ tuple,  $P(s'|s,a)=1$  
for exactly one $s'$. If the domain is partially observable, the state observation 
mapping is also assumed to be deterministic but may involve aliasing of two states 
having the same observation, e.g. $p(z_1|s_1,a_1)=p(z_1|s_3,a_1)$. This implies 
that executing an action from a given state will yield a single next observation. 
The dynamics model is assumed to be unknown to the learning agent.

Note that the learning agent's observation space must include the set 
of distinct observations in the observation demonstration $\mathcal{Z}^*$ but 
the action space of the learning agent may not match the demonstrator's 
action space. For example, a series of photos may show the steps of creating an omelette by a human chef, but a robot could learn to perform the same task and generate the same photos.


In many situations the observed procedure may itself consist of multiple subtasks which 
can repeat multiple times within a task. 
Similar to the policy sketch notation~\cite{Andreas2017ModularMR} we assume that there is an underlying procedure sketch 
$K^*=(b_{1},b_{2},\ldots,b_{L})$ where each element of the sketch is a label for a particular open-loop action sequence drawn from a fixed set $\mathcal{B}$, departing slightly from the original policy sketches work in which each subtask was a policy. The actual action sequence 
associated with each element is unknown. An example of these subtasks in one of our domains is shown in Figure~\ref{robot_piano}.

\section{ALGORITHM}
We present two versions of our online procedure learning algorithm\footnote{All code \url{https://github.com/StanfordAI4HI/PLOTS}}:
\begin{enumerate}
    \item \textbf{\algi}~is given the task sketch and uses it to infer subtask assignments and alignments.
    \item \textbf{\algii}~is not given the task sketch and instead infers and stores possible low level action sequences that could potentially be subtasks.
\end{enumerate}

There are two main insights to our approach. The first is to leverage the deterministic structure of the procedure imitation setting to systematically search for a sequence of actions that will enable the learner to match the desired observation trajectory. The second is to strategically use the potential presence of repeated structure to guide exploration. 

\subsection{Procedure Imitation As Structured Search} \label{subsec:baseagent}

Recall the agent's goal is to learn how to imitate a fixed input 
sequence of observations $\mathcal{Z}^*=z^*_1,z^*_2,\ldots,z^*_H$. For this to be possible 
we assume that the dynamics of the underlying domain is deterministic, 
at least in terms of the actions available to the agent in order to 
achieve the desired observation sequence. Note that we \emph{do not} assume that the 
observation space is necessarily Markov. 

Our algorithm proceeds by incrementally learning a sequence of actions that yields the observation sequence $\mathcal{Z}^*$. 
Notice that $\mathcal{Z}^*$ provides dense labels/rewards after an action $a_t$ taken at time
step $t$, since the agent sees its next observation $\tilde{z}_t$ and 
can immediately identify if $\tilde{z}_t$ matches the desired observation
$z^*_t$. If it matches, then $a_t$ is identified as a candidate for the correct action at time step $t$ and is added to a partial solution action trajectory 
$a^*_{1:t}$. 
The agent then continues, trying a new action $a_{t+1}$ to match $z^*_{t+1}$.

If $\tilde{z}_t$ \emph{does not match} the 
desired observation $z^*_t$, the agent simply plays random actions until the end of 
the trajectory $H$. It is then reset to the start state, and follows the  
known partial solution action trajectory $a^*_{1:t-1}$ until it reaches time step 
$t$, and then with uniform probability chooses an action that has not yet been tried for $t$.

In general, aliasing may occur if the observation space is not 
Markov. In such cases, even if an action $a_t$ yields the 
desired observation $z^*_t$, the latent state underlying the agent's observation $z_t$ may be wrong due to aliasing, preventing the agent from mimicking the rest of the sequence. 
This is detectable when an agent reaches a later time step $t'$ for which 
no actions can yield the specified observation $z^*_{t'}$. In this 
case the agent backtracks $a^*_{1:t-1}$ one step, $a^*_{1:t'-2}$ and restarts the process from there to find new 
actions that yield the same remaining procedure observation sequence, possibly backtracking again when necessary. We will refer to the agent that does this as {\algbaselong}.

\noindent\textbf{Learning Efficiency of \algbase.}
If the observation space is Markov given the agent's actions, then 
once an action at time step $t$ yields the specified desired 
next observation $z_t$, that action never needs to be revised. For a Markov 
state at most $|A|$ actions must be explored. Since each "failed" action 
attempt requires the agent to act until the end of the episode and 
then replay the learned solution action sequence up to the desired 
time step $t$, it can take at most $H|A|$ time steps for the agent 
to learn the right action to take in time step $t$. Repeating this 
for all $H$ time steps yields a total sample complexity of $|A|H^2$ 
to learn the procedure. This matches the expected sample 
complexity for deterministic tabular Markov decision processes, 
since only a single sample is needed to learn each state--action 
dynamics model.  Note that if we were to treat this 
problem as a policy search problem, the number of possible policies 
is $|A|^{|S|}$ or $|A|^H$ if each observation is unique in the procedure 
demonstration. In general this will be substantially less efficient 
than our method. 

If the observation space is not Markov and aliasing occurs, 
 in the worst case, the process of backtracking and going forward may occur repeatedly until all $|A|^H$ possible action trajectories 
 are explored. This matches the potential set of policies considered 
 by direct policy search algorithms for this domain, that are also
 robust to non-Markovian structure. However, in practice we 
 rarely encounter such cases, and we find that
 our approach only has to perform infrequent backtracking.
 
\subsection{Exploration using Sequence Substructure}
The \algbase~ algorithm described above is agnostic to and does not utilize the presence of any hierarchical structure
.
To leverage potential repeated subsequence structure, we extend \algbase~ by proposing the {\sc PLOTS}s -- which provide heuristics for action selection resulting in smarter exploration. Note that accounting for repeated structure should provide significant speedups if such structure exists, but if no such structure exists then the {\sc PLOTS} algorithms should perform equivalently to \algbase.

\subsubsection{\textbf{\algi}}
\noindent For \algi, in addition to maintaining a search tree to build a potential solution action trajectory $a^*_{1:H}$, our algorithm also maintains a finite set of partial action sketch instantiation hypotheses. As a concrete example, consider the observed procedural sequence ($z_1$, $z_2$, $z_3$, $z_4$, $z_5$, $z_6$, $z_7$, $z_8$) and the associated subtask sketch ($b^1,b^2,b^1,b^3,b^1$). Let the agent have learned that the first 5 actions 
are $a^*_{1:5}=(e,f,g,e,f)$. Then two potential partial action sketch assignments are $\tilde{M}_1=[b^1={e,f},b^2={g}]$ $\tilde{M}_2=[b^1={e},b^2={f,g}]$
Both of these partial action sketch hypotheses are consistent with the learned partial solution action trajectory $a^*_{1:5}.$ Yet they have different implications for the optimal action sequence in the remainder of the trajectory.

The two primary functions we must address is how to use potential action sketch hypotheses to facilitate faster learning, and how to update existing and instantiate new hypotheses. 

\noindent{\textbf{Action Selection Using Partial Action Sketch Hypotheses}}. 


\noindent To use these partial action sketch hypotheses for action selection, at each timestep $t$, all hypothesis tracked by the agent can potentially suggest an action to take next using the following guidelines:

\begin{itemize}
    \item If the hypothesis estimates the current time step $t$ is in a subtask for which it has an assignment, it will execute the next action in that subtask. 
    \item Otherwise, the hypothesis returns $NULL$ to the agent, indicating that it does not have any action suggestions.
\end{itemize}
In practice we found a slight variant of the above score function and action selection procedure was beneficial. Instead of returning $NULL$, the hypothesis 
makes an optimistic assumption that the first repeating subtask that has not yet been assigned will repeat as soon as possible and will have length as long as possible.
For example, consider at timestep $t=4$ a new hypothesis $\tilde{M}_3 = []$, which has not yet instantiated any potential mappings of subtasks to actions. At $t=4$ the partial action solution is known to be ${e,f,g,e}$. The first repeated subtask in this case is known to be $b^1$ and it is also known that $b^1$ aligns with the beginning of the partial action solution. Due to the non-emptiness of subtasks, we know the first $e$ found at $t = 1$ in $a^*_{1:5}$ belongs to $b^1$. So we optimistically assume that $b^1$ is  currently repeating and the second $e$ found at $t=4$ is the result of $b^1$ repeating as opposed to belonging to $b^2$. We also optimistically assume $b^1$ is as long as possible and the $f$ found at $t=2$ also belongs to $b^1$ as opposed to $b^2$. With these optimistic assumptions, the next action should be $f$ which $\tilde{M}_3$ will suggest instead of suggesting $NULL$.

With each hypothesis possibly suggesting an action, the agent must select a hypothesis to follow. To this end, we compute a score for each hypothesis and use this score to select among them.
The score $C(\tilde{M}_i,t)$ is the maximum reduction in time needed to learn the remaining procedure that could result if that hypothesis $\tilde{M}_i$ were true and is calculated as:
\begin{equation}
\label{eqn:score}
    C(\tilde{M}_i,t)=\sum_{j=1}^L N_{b_j} l(\tilde{M}_i(b_j)),
\end{equation}
where $N_{b_j}$ is the number of repeats of subtask $b_j$ in the remainder of the procedure given hypothesis $\tilde{M}_i$, and $l(\tilde{M}_i(b_j))$ is the length of the action subsequence cooresponding to subtask $b_j$ in $\tilde{M}_i$. Note that if $\tilde{M}_i$ does not include a hypothesized assignment for element $b_j$, then its length is assigned to be 0. Continuing our running example, consider computing the score for $\tilde{M}_1$ after $t=5$. Under this hypothesis, the remaining sketch for the rest of the trajectory is only $(b^3,b^1)$ since $\tilde{M}_1$ hypothesizes that $(b^1,b^2,b^1)$ have already been observed. Therefore, 
\begin{equation}
\begin{split}
    C(\tilde{M}_1,t=5) & =\sum_{j=}^L N_{b_j} l(\tilde{M}_i(b_j)) \\
    & = N(b^1) l({e,f}) + N(b^3) l(\null) = 1*2 = 2
\end{split}
\end{equation}
since $\tilde{M}_1$ does not include an instantiation for $b^3$ so $l(\tilde{M}_1(b^3))=0$ and $b^1={e,f}$ under this hypothesis. 

To use this score to select a hypothesis, recall that in discrete domains, at each timestep $t$ the agent learns the correct action by trying actions until the correct one is found and the observed next state $\tilde{z}_{t+1}$ matches the correct state at $t+1$, $z^*_{t+1}$. Let $A'_t$ be the set of all incorrect actions the agent has tried at $t$. Let $\mathcal{H}_{A', t}$ represent the set of hypotheses tracked by the agent at time $t$ that are not suggesting an action in $A'_t$ and that are not suggesting $NULL$. After all scores are computed for the tracked hypothesis, the partial sketch $\tilde{M}*=\arg\max_{\tilde{M}_i \in \mathcal{H}_{A', t}} C(\tilde{M}_i,t)$ with the highest score is selected. The agent then follows the action suggested by this hypothesis.

\noindent{\textbf{Hypothesis Creation and Updating}}.
Whenever the agent reaches a time step $t$ on which it adds a new partial solution action trajectory element $a_t$ that is a repeat of a previously encountered action in the current solution trajectory, new subtask action hypotheses can be introduced. To reduce computational complexity, the agent only reasons about assignments for one subtask at a time and additional subtasks get assigned only if the assignment of the main subtask immediately implies it. 
To reduce the memory complexity of enumerating and storing all possibilities, we only create hypotheses for subtasks assignments we have \textit{consistent evidence} to be true in the sense that we have seen a consistent alignment where that subtask assignment has already repeated at least one. 
To continue with our example, consider again the timestep $t = 4$ and $\tilde{M}_3$, the hypothesis which has not yet instantiated any mappings. For this hypothesis, the first new item is $b^1$ so the main hypothesis it is trying to find an assignment for is $b^1$. At $t = 4$, the partial action solution is $a^*_{1:5} = {e,f,g,e}$, and from $\tilde{M}_3$ the agent can instantiate $\tilde{M}_2=[b^1={e},b^2={f,g}]$ because we have consistent evidence in $a^*$ of $b^1 = {e}$, meaning, we have seen ${e}$ repeat at least once in $a^*$ and assigning $b^1 = {e}$ is consistent with the assumptions made about the subtask structure. By assigning $b^1 = {e}$, it immediately applies $b^2 = {f, g}$ in this hypothesis so we also make an assignment for $b^2$. However, we do not instantiate $\tilde{M}_1 = [b^1={e, f},b^2={g}]$ or other hypotheses that would assign $b^1 = {e, f, g}$ or $b^1 = {e, f, g, e}$, etc. because at $t=4$, we have not yet seen those sequences for $b^1$ repeating in a consistent manner. We also continue to track $\tilde{M}_3$ which has not yet instantiated any mappings but we will name it $\tilde{M}_4$ for clarity. Now consider moving forward to the next timestep after discovering the next correct action is $f$. Now the agent is at $t=5$, and $a^*_{1:6} = {e, f, g, e, f}$. At this point from $\tilde{M}_4$, we can branch and instantiate $\tilde{M}_2 = [b^1={e, f},b^2={g}]$ because we have now seen the sequence ${e, f}$ repeat in a consistent manner. 

From this example, we can notice that we only need to find instantiations for the main hypothesis where the repeats match at the end of $a^*$. For example at $t=5$, $\tilde{M}_4$ even though we have consistent evidence for $b^1 = e$, we do not need to re-instantiate that because we have already instantiated that hypothesis at $t=4$.

\noindent{\textbf{Computational Tractability}}. 
Like in beam search, for computational tractability we maintain only a finite set of subtask action hypotheses. As previously mentioned, whenever the agent finds a new partial solution action element $a_t$ for a time step $t$, new subtask action hypotheses can be introduced.
Each existing hypothesis can generate at most $H/2$ new hypotheses on a given time step $t$. To see this, we deviate from our running example and present a new example. Consider the situation where the subtask sequence is ${b^3, b^1, b^2, b^1, ...}$ and the agent is at timestep $t = 8$ with a $a^*_{1:8} = {e, f, g, h, i, f, g, h} $. Let one of the hypothesis the agent is tracking be $\tilde{M}_5 = []$, one that has no hypothesized subtask assignments. At this timestep, we instantiate the following assignments for $b^1$ (and by immediate implication also make assignments for $b^2$ and $b^3$) all of which we have consistent evidence for: $\tilde{M}_6 = [b^1={ h},b^2={i}, b^3 = {e, f, g}]$, $\tilde{M}_7 = [b^1={g, h},b^2={i}, b^3 = {e, f}]$, $\tilde{M}_6 = [b^1={ f, g, h},b^2={i}, b^3 = {e}]$.
Because we only instantiate a hypothesis once we see repeats, the greatest number of branching we can have at each step is at most $H/2$.
Though each individual hypothesis will only generate at most a polynomial number of additional hypotheses at each time step, repeating this across many time steps can yield an exponential growth. Therefore we maintain a finite set of $N_1$ potential hypotheses which we actively update and we do not the update the rest. This is done via two mechanisms. First, hypotheses are ranked according to the score function (Equation~\ref{eqn:score}) and only the top $N_1$ are kept active. We will refer to the hypotheses not in the top $N_1$ that we are not tracking as frozen. Second, if the current hypothesis is inconsistent with the observed procedure and partial action solution trajectory, that hypothesis is eliminated. This can occur later during the procedure learning when additional discoveries of elements of $a^*$ make it clear that an earlier hypothesis is inconsistent. To maintain the correctness of our algorithm, if we reach a point in where we have no more tracked consistent hypotheses, we can unfreeze frozen hypotheses and continue. Empirically, we have found that in the domains we considered, our sorting metric works well and if a reasonable number of hypotheses are tracked, then very little unfreezing needs to be done. This leads to a memory complexity of $O(H^2)$ in terms of the number of hypotheses stored. Pseudocode for \algi~ is presented in Alg~\ref{alg:PLOTS}. 

\begin{algorithm}
 \caption{\algi} \label{alg:PLOTS}
 \begin{algorithmic}[1]
 \State $d$ (\# hypotheses to track), $\mathcal{Z}^*$ (observation sequence)
 \State $\mathcal{M} = \emptyset$, $a^* = \null$ // actions yielding partial match of $\mathcal{Z}^*$
 \State $\mathcal{A}_p = \{1:|A|\}$, $i=1$ // episode number 
 \While{$|a^*| < H$} // haven't learned full procedure
 \State Reset to $s_0$, $t=0$
 \State Execute $a^*$ // execute known subprocedure
 \State $t=|a^*|+1$, 
 \State Evaluate score $ C(\tilde{M},t)$ for each hypothesis $\mathcal{M}_a$
 \State $a_t \leftarrow$ Action from $\arg\max_{\tilde{M}}  C(\tilde{M},t)$
 \State Execute $a_t$, observe $z_{i,t+1}$
 \If{$z_{i,t+1} == z^*_{t+1}$} 
 \State // Found action that yields observation
 \State $\mathcal{M} \leftarrow$ UpdateActiveH
 \State $a^*$ $\leftarrow$ ($a^*$,$a_t$)
 \ElsIf{$\mathcal{M} == \emptyset$}
 \State No consistent active hypotheses
 \State Backtrack to unroll past incorrect actions \& reset $\mathcal{M}$ 
 \EndIf
 \EndWhile
    \end{algorithmic}
\end{algorithm}



\subsubsection{\textbf{\algii}}
\noindent The \algii~ algorithm is not given the task sketch and relies on the fact that the task consists of repeating subtasks. At each timestep, \algii~ looks into the partial solution action trajectory $a^*_{1:t}$ for repeated sequences of low level actions. Repeated low level action sequences, or hypothesized subtasks, are stored along with the number of times they are repeated. To reduce the computational complexity of this method, we only add and update the counts of repeated action sequences that also match at the end of $a^*_{1:t}$. This is sufficient because other repeated action sequences will have been discovered and updated at previous time steps. To suggest an action, we sort all hypothesized subtasks by the number of times they have repeated. We then follow in that order the consistent next actions of hypothesized subtasks until the correct action for time $t$ is found.



\section{EXPERIMENTS}

\begin{figure}[t!]
    \centering
    \begin{subfigure}[t]{0.14\textwidth}
        \centering
        \includegraphics[width=\textwidth]{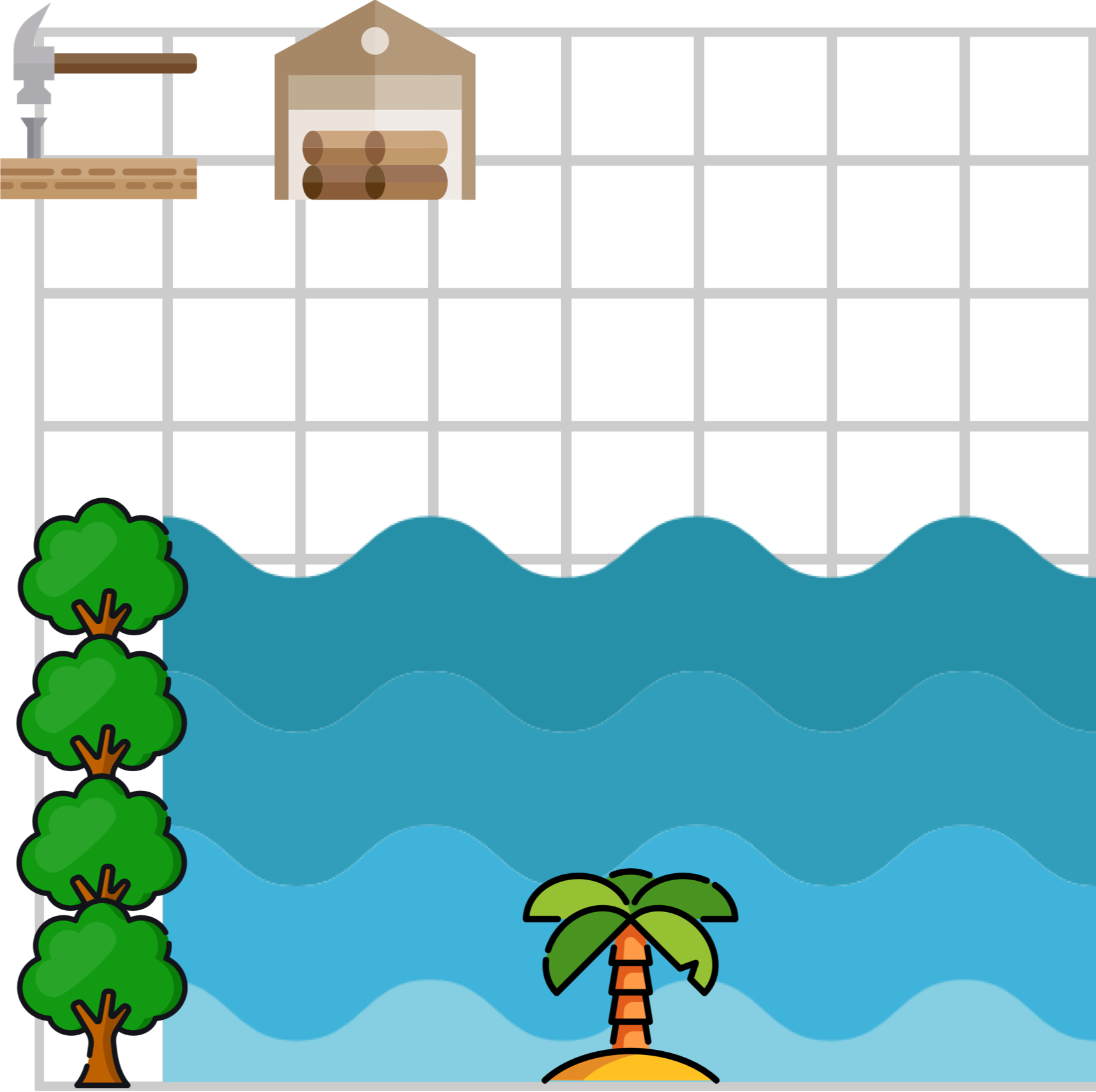}
  \caption{Island}
  \label{fig:craftisland_world}
    \end{subfigure}
    \begin{subfigure}[t]{0.14\textwidth}
        \centering
        \includegraphics[width=\textwidth]{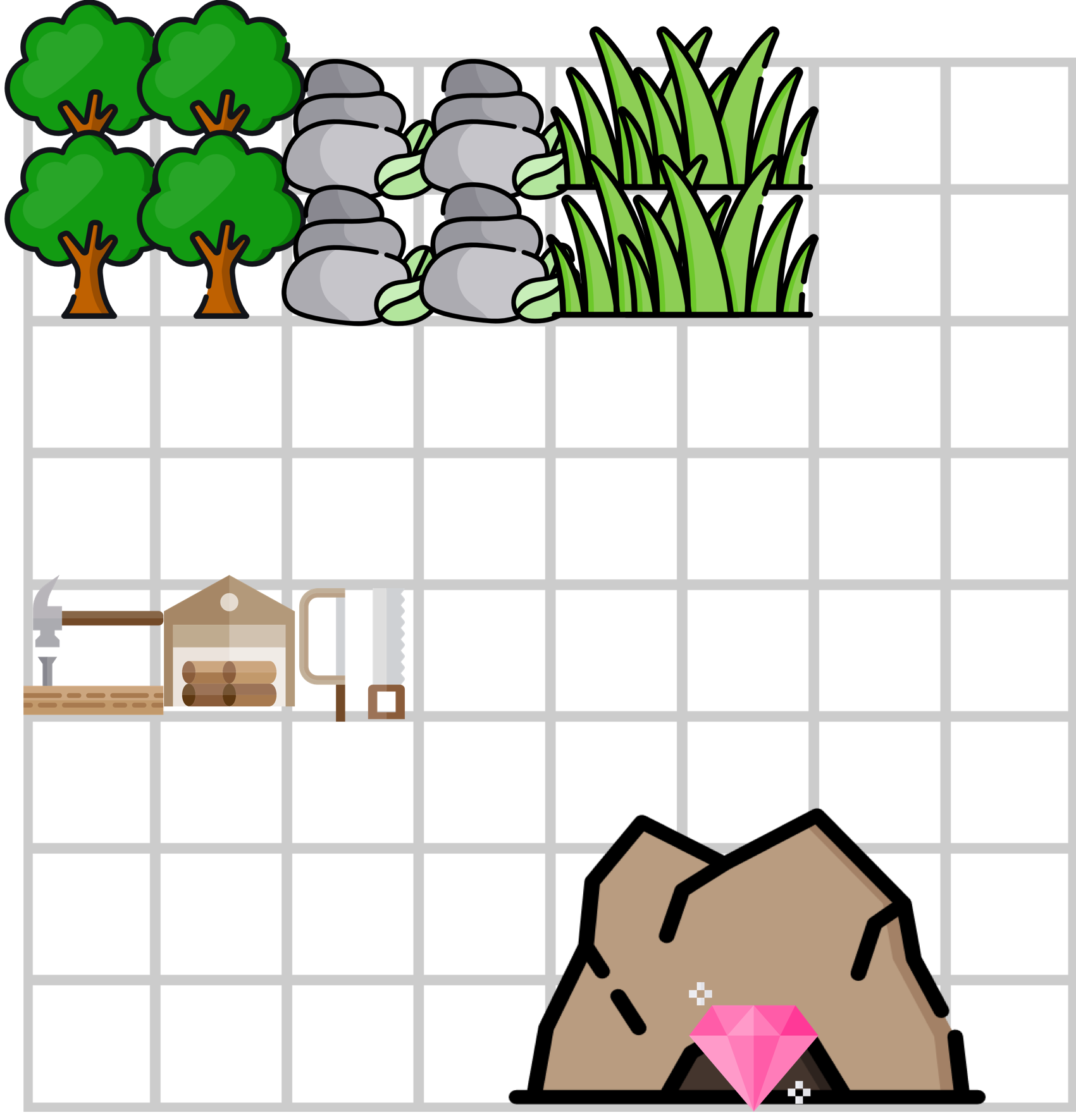}
  \caption{Gem}
  \label{fig:gem_world}
    \end{subfigure}
    \begin{subfigure}[t]{0.12\textwidth}
        \centering
        \includegraphics[width=\textwidth]{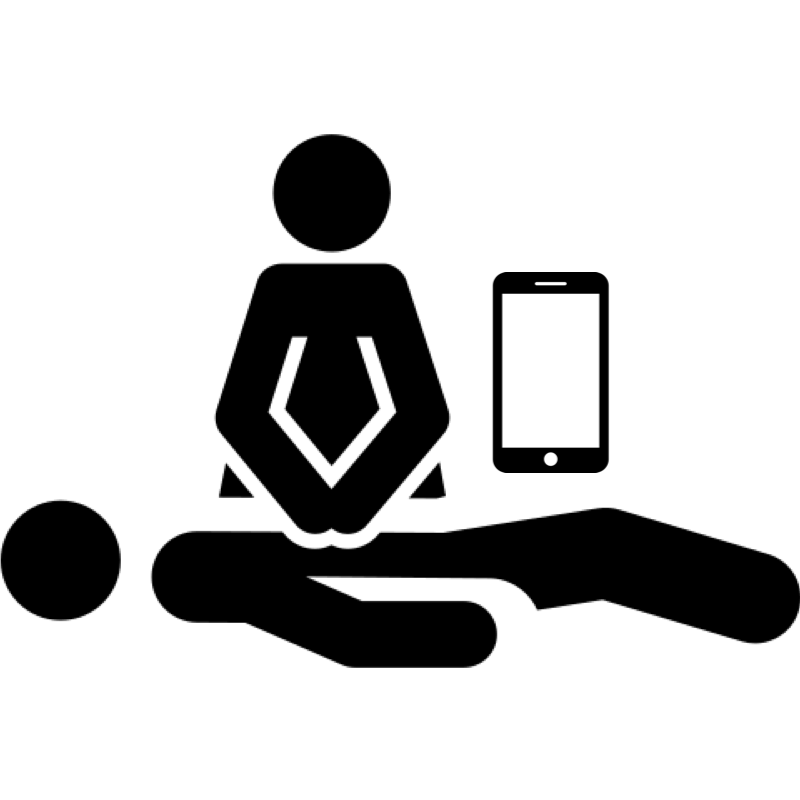}
  \caption{CPR}
  \label{fig:CPR_world}
    \end{subfigure}
    \caption{Illustrations for the domains~\cite{image_credits}}.
\end{figure}

    

We compare our methods against state-of-the-art baselines that can learn observational procedures and baseline versions of our method.
Our methods are summarized below
\begin{enumerate}
    \item \textbf{\algi}~ is given the procedure sketch and leverages it to hypothesize about the assignments of low level actions to subtasks and the alignment of the procedure sketch to the state sequence to perform smarter exploration.
    \item \textbf{\algii}~ is not given the procedure sketch. It leverages the fact that the task is made of repeated subtasks and hypothesizes possible action sequences that could correspond to subtasks to use for smarter exploration.
\end{enumerate}
We compare with baseline version of our method:
\begin{enumerate}
    \item \textbf{\algbase}~ is described in section~\ref{subsec:baseagent}. It is not given the procedure sketch and does not infer or leverage any of the repeated hierarchical structure.
    \item \textbf{\algoraclelong}~ is given the oracle alignment of the procedure sketch to the state sequence in addition to the procedure sketch. This agent is able to learn faster because it does not need to hypothesize about the alignments and only needs to learn the assignment of action sequences to subtasks.
\end{enumerate}

\noindent{We compare against state-of-the-art policy gradient based methods\footnote{Code for \textsc{GAIL} which we additionally modified to obtain our \textsc{Policy Gradient} baseline is taken from \url{github.com/openai/baselines} and \textsc{Modular} from \url{github.com/jacobandreas/psketch}.}}
\begin{enumerate}
    \item \textbf{\textsc{Modular}} \citep{Andreas2017ModularMR} leverages the sketch to learn the procedure. Originally {\sc Modular} was used to learn multiple tasks with sparse rewards using the sketches, but we instead provide the method with dense per-step rewards for our setting.
    \item \textbf{\textsc{Gail}} \citep{Ho2016GenerativeAI} is an imitation learning method that learns to imitate the given observation sequence by adversarial training. This method is not able to leverage the sketch.
    \item \textbf{\textsc{Policy Gradient}} (PG) is adapted from \textsc{Gail} \cite{Ho2016GenerativeAI} and replaces the discriminator with per-step rewards to result in a purely policy-gradient approach. We reason that this could potentially be more efficient than \textsc{Gail} as instead of learning the reward function (discriminator) we directly provide it. This method is also not able to leverage the sketch.
\end{enumerate}


These baseline methods all rely on a policy-gradient approach to learn a stochastic policy, rather than learning an open-loop plan like \alg~ variants and baselines. Since our procedure is deterministic and our methods are specialized to learning in deterministic domains where open-loop plans are sufficient, we expect these baselines will all converge more slowly to a locally optimum policy. However, they do have the additional benefit of being able to leverage a deep neural network to internally learn a state abstraction.  Because many deep neural network approaches are sensitive to hyperparameters, for the results reported for each of the baselines, we did a basic hyperparameter sweep over 4-6 different sets of hyperparameters and display the set that performed best.

We also compare against model-based methods which we modify to be computationally tractable in our domains which have large state spaces. As with the policy gradient based approaches we also provide dense, one step rewards signaling whether the agent has found the correct action to perform the procedure.
 \begin{enumerate}
    \item \textbf{\textsc{RMax+}} \citep{brafman2002r} a tabular model based algorithm that initializes the values of all states optimistically. For computational tractability, we build up the Q-value, reward, and transition tables as we see new states and group all states that were not on the demonstration trajectory as the termination state. 
    \item \textbf{\textsc{UCB+}} \citep{auer2002finite} a bandit algorithm that keeps track of confidence intervals of the rewards of the arms and chooses the arm with the highest upper confidence reward. We apply this by treating each unique state as a separate bandit problem. Because we are only considering the deterministic case, the exact reward of a state action pair $(s,a)$ can be learned after one attempt of the action in the state and the confidence interval shrinks to zero. For tractability we also build up the number of bandit problems as we see new states and treat all states that were not on the demonstration trajectory as the degenerate bandit where all actions lead to zero rewards. 
\end{enumerate}
 
 Note that we do not compare to methods that require the demonstrator's actions to be provided, such as behavior cloning and recent variations on this \citep{shiarlis2018taco,Fox2017MultiLevelDO}, since we assume we do not have or are not able to utilize the demonstrator's actions.

\subsection{Environments\protect\footnote{All code and more detailed environment descriptions https://github.com/StanfordAI4HI/PLOTS}}

\subsubsection{Craft Domain}
A discrete 2D-domain introduced by Andres et al \citep{Andreas2017ModularMR} to evaluate policy learning using policy sketches in multitask domains with sparse rewards. In this domain, the agent is required to complete various tasks by moving and interacting with objects using 5 deterministic actions: up, down, left, right, use. 
The tasks have hierarchical structure so each task has a corresponding policy sketch. This domain was first proposed to demonstrate the effectiveness of an algorithm that learned policies in a multitask setting. Therefore in the original tasks, a single task did not have any repeated subtasks but the agent could leverage repeated subtasks across multiple tasks to speed learning. This differs from our setting, since we are primarily interested in the single task setting where there is repeated structure within a task. Therefore to evaluate our method, we create a new task that involves collecting multiple wood objects, forming them into planks, and using them to building a raft to reach an island (Island, Fig~\ref{fig:craftisland_world}). This procedure is length $H=67$ with a policy sketch of length $L=16$ consisting of $|\mathcal{B}|=9$ unique subtasks. 
Additionally, we also use one of the original tasks from this domain (Gem, Fig~\ref{fig:gem_world}) which does not have repeated subtasks in the task sequence, to evaluate the benefit our method obtains from being tailored to deterministic domains. 

\begin{figure*}[t!]
    \centering
    
    \begin{subfigure}[t]{0.3\textwidth}
        \centering
         \includegraphics[width=\textwidth]{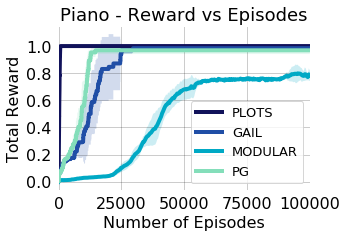}
  \caption{}
  \label{fig:piano}
    \end{subfigure}
    \hfill
    \begin{subfigure}[t]{0.3\textwidth}
        \centering
        \includegraphics[width=\textwidth]{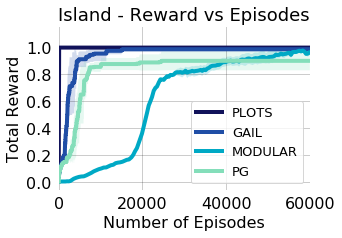}
  \caption{}
  \label{fig:craft3}
    \end{subfigure}
    \hfill
    \begin{subfigure}[t]{0.3\textwidth}
        \centering
        \includegraphics[width=\textwidth]{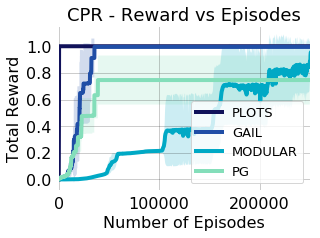}
  \caption{}
  \label{fig:cpr}
    \end{subfigure}
    \hfill
    \begin{subfigure}[t]{0.3\textwidth}
        \centering
        \includegraphics[width=\textwidth]{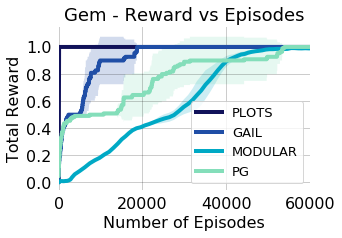}
  \caption{}
  \label{fig:craftgem}
    \end{subfigure}
    \hfill
    \begin{subfigure}[t]{0.3\textwidth}
        \centering
        \includegraphics[width=\textwidth]{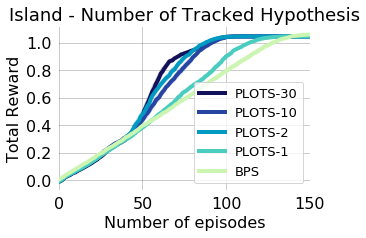}
  \caption{}
  \label{fig:craftsweep}
    \end{subfigure}
    \hfill
    \begin{subfigure}[t]{0.3\textwidth}
        \centering
        \includegraphics[width=\textwidth]{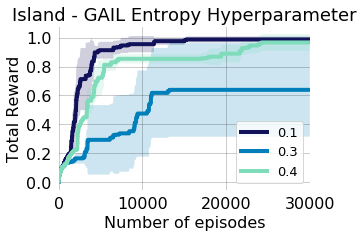}
  \caption{}
  \label{fig:craftsweepgail}
    \end{subfigure}
    
  \caption{Comparing PLOTS with policy-gradient based baselines in four discrete domains (a) Piano (b) Island (c) CPR (d) Gem. A basic hyperparameter sweep was done for the baselines and the best set of hyperparameters were chosen. The Gem domain (d) did not have any repeated structure and shows the speedup of our method that is specialized to learning procedures over using more general procedures. Our approach was able to learn 10 - 100 orders of magnitude faster. For the Island domain, we also show a hyperparameter sweep of our algorithm (e) and for {\sc \textbf{GAIL}} (f). In all plots, {\sc \textbf{PLOTS}}~ refers \algi~ and for (e), the -number refers to the number of hypothesis tracked.}
\end{figure*}
\subsubsection{CPR Domain}
(Fig~\ref{fig:CPR_world}) The task of the agent in CPR world is to follow the correct steps necessary to perform CPR on a patient based on standard CPR procedures\footnote{https://www.redcross.org/take-a-class/cpr/performing-cpr/cpr-steps.}. The agent has 23 actions 
that are used in the observation demonstration of length $H=197$, with a policy sketch of length $L=6$ consisting of $\mathcal{B}=2$ unique subtasks. 


\subsubsection{Piano Domain}
In the Piano domain an agent learns to play the right hand component of Bach's Prelude in C (boxed in blue in Fig~\ref{robot_piano}) in a simulated piano environment. The observation sequence has $H=64$ notes, with a policy sketch of length $L=24$ consisting of $|\mathcal{B}|=5$ distinct subtasks 
The agent has a 5 fingered manipulator and the action space is to press each of 5 fingers down, move the whole wrist up one note, move the whole wrist down one note, or move only the thumb down or up one note (with a max range of 3). This yields a total of 9 actions. The observation space is the audio of the note and not the hand position. This yields a partially observable state space since multiple hand positions can be used to play the same note.

\begin{table}[b!]
    \centering
    \resizebox{\columnwidth}{!}{%
    \begin{tabular}{c|c|c|c|c|c|c|c|}
        {\sc Env} & \makecell{\sc PLOTS-\\\sc Sketch} & \makecell{\sc PLOTS-\\\sc NoSketch}  & \algbase & \algoracle & \sc RMax+ & \sc UCB+ & \sc GAIL\\\hline
        {\sc Island} & 92 & 80 & 137 & 72 & 265 & 265 & 20198\\
        {\sc Gem} & 43 & 43  & 44 & 43 & 86 & 86 & 9892\\
        {\sc Piano} & 405 & 392 & 539 & 206 & 30,000+ & 30,000+ & 18526\\
        {\sc CPR} & 319 & 286 &  2005 & 455 & 4230 & 4302 & 25653\\
        \hline
    \end{tabular}%
    }
    \caption{Average number of episodes until the procedure is learned for \algi~ and baselines, only {\sc GAIL} is listed amongst the policy gradient based baselines as it did best.}
    \label{tab:PLOTS_results}
\end{table}

\subsection{Benefits of Procedure Learning}
Figures~\ref{fig:piano}, \ref{fig:craft3}, \ref{fig:craftgem}, \ref{fig:cpr} and Table~\ref{tab:PLOTS_results} display the results of 
running our approach and baselines on the Craftworlds, Piano and CPR simulation 
domains. From the figures, in all cases we observe that our procedural learning from 
observation action requires at least 100 times less 
episodes to learn the desired procedure than the baseline 
policy learning algorithms. This clearly illustrates the enormous 
benefit of leveraging knowledge of the deterministic dynamics in order to incrementally compute a plan. Note this is true both 
in the large state space Markov domains (Craftworlds) as well 
as the partially observable Markov domain (Piano). 

Additionally we can see from Table ~\ref{tab:PLOTS_results} all our algorithms performed significantly better than the model-based baselines. This improvement results from our method not optimistically exploring all possibilities but instead focusing on finding a single plan that achieves the desired full sequence. Additionally, model-based baselines do not perform well in the Piano domain where a Markov model is  history-dependent and requires exploration over an exponential history space of $O(|A|^H)$.


It has been recently 
observed that curriculum learning can speed reinforcement learning, 
and indeed the policy sketches algorithm employed hand-designed 
curriculum learning across different length sketches during their 
multi-task training procedure~\cite{Andreas2017ModularMR}. One 
might wonder if curriculum learning could be applied to improve 
the performance of the baselines in these domains, since our 
own approaches implicitly perform incremental curriculum learning 
as they slowly build up a correct action plan that yields the 
desired observation sequence. To mimic this process, one could 
imagine first training a policy network to first correctly obtain the 
first observation, then train 
it to correctly obtain the first two observations, etc. Unfortunately, 
in partially observable environments, at some point it is likely 
that the previously trained policy for an earlier observation is 
incorrect. In our approaches this is where systematic backtracking can be done, 
to efficiently unroll/unlearn proposed solution action plans. 
However, in generic policy training, this additional guidance about 
how to start searching for alternate policies, and which parts to 
revisit, is entirely unstructured, making it likely that this could 
incur a general cost of expanding all prior $|A|^H$ decisions. In contrast, 
our method typically only backtracks a small number of times, yielding a 
final computational cost that is closest to a linear $C_3$ scale up of the 
Markovian decision space $C_3 |A|H$ rather than needing to explore the 
full exponential space.

\subsection{Utilizing Substructure Can Speed Learning}
Table~\ref{tab:PLOTS_results} additionally shows a comparison between the variants of our method, against our own baselines,
illustrating that our algorithm 
variants that 
leverage knowledge of the 
subtask structure within the observation demonstration learn with substantially less 
episodes than our variant, \algbase, which is agnostic to potential substructure.
The Gem example which has no repeated action substructure illustrates that if no substructure exists, all of our algorithms perform similarly, as expected.

Interestingly, note that sometimes our algorithms that do not 
receive the ground truth alignment outperform the oracle 
variant, $\algoracle$. We find that in practice there may be 
repeated action subsequences that can't yet be confidently 
aligned with particular observations, but that optimistically 
assuming such alignments can yield substantial speedups.
Indeed, in many of the problems, there is additional 
substructure that is not reflected in the sketches. For example, 
in Island, one open loop action subsequence 
could be to travel from the workshop to the forest entrance 
(the place we term that is around all the wood)
using a action sequence that has one action repeated many times (for example Down, Down, Down, Down, Down, Down, Left). In this case there is additional substructure, (Down, Down, Down), that \algii~ is able to use that can allow it to 
perform better than \algi. However this result is specific to the
problem structure where there is additional substructure within 
a subtask open loop plan. 

The above experiments illustrate the benefit of action substructure. 
To better understand the potential impact on agent learning of 
strategic action substructure hypothesis generation to inform 
action selection, we explored the sensitivity of the \algi~ 
algorithm to the number of tracked hypotheses, our main hyperparameter(Figure\ref{fig:craftsweep}). 
We find a significant jump from using at least 2 hypotheses, but 
more yield minor differences. This illustrates that being able to 
strategically suggest potentially beneficial actions given a small 
set of hypotheses can be beneficial and computationally tractable (due 
to the low number of tracked hypotheses).

\section{DISCUSSION AND FUTURE WORK}
Our experiments show that \algi~ and \algii~ are capable of quickly learning a given procedure, leveraging the sketch to discover macro-actions that can be reused later on. Our results show that for learning procedures with deterministic dynamics, specialized algorithms for learning procedures with deterministic dynamics, focusing on specialized algorithms can be vastly more efficient than more general policy-gradient style methods which are able additionally able to learn stochastic policies. In this work we focus on discrete domains since many domains are naturally discrete or near discrete. We have preliminary work in successfully adapting our algorithm to domains with both continuous state and action spaces, using gradient descent on the action space in domains where the reward is continuous and convex with respect to the action. Note that in continuous state spaces, it is impossible to match the observation state exactly. Thus, we approximately match observations, with a tolerance on the $l_2$ distance between the agent and demonstrator observations. Due to this approximation, we cannot directly apply the learned actions of a subtask as-is, due to compounding errors, however learning subtask assignments is still useful in that they provide a favorable initialization for the action search, allowing the number of episodes needed to find an approximately correct action to be half of the number episodes needed with a random initialization when using some gradient based optimization methods such as COBYLA \cite{powell1994direct}. Additionally in this work we do not consider stochastic domains; in such domains, without additional assumptions, it is impossible for any algorithm to guarantee that it can find a policy or action sequence to match the observed procedure. However an area of future exploration is stochastic domains where the dynamics appear deterministic given an appropriate state abstraction \cite{zhang2018composable}. 


\section{CONCLUSION}
We introduce \algi~ and \algii, novel approaches for learning to imitate deterministic procedures in tasks that have repeated structure in the form of subtasks. \algi~ is able to incorporate additional information in the form of a procedure sketch to help reason about action to subtask assignments and speed learning. \algii~ inferred possible action sequences that could correspond to subtasks without the sketch information.  We evaluated the performance of our algorithms in four different domains, including a domain that is partially observable in the state space. Our algorithm for learning procedures in discrete deterministic domains vastly outperformed related methods designed for general classes of problems.

\section*{ACKNOWLEDGEMENTS}
This material is based upon work supported by the Schmidt Foundation, the NSF CAREER award and the National Physical Science Consortium fellowship.

\end{document}